\documentclass{article}

\usepackage{arxiv}

\usepackage[utf8]{inputenc} 
\usepackage[T1]{fontenc}    
\usepackage{hyperref}       
\usepackage{url}            
\usepackage{booktabs}       
\usepackage{amsfonts}       
\usepackage{nicefrac}       
\usepackage{microtype}      
\usepackage{lipsum}
\usepackage{graphicx}
\graphicspath{ {./images/} }

\title{Impact of Sentiment Analysis On Fake Review Detection

Literature Survey}

\author{
 Amira Yousif  \\
 Department of Computer Science\\
  University of Dayton \\
  Dayton, OH 45469\\
  \texttt{ayousif1@udayton.edu} \\
     \And
 James Buckley\\
Department of Computer Science\\
  University of Dayton\\
  Dayton, OH 45469\\
  \texttt{jbuckley1@udayton.edu} \\
}
 \usepackage{setspace}
\begin{document}
\maketitle
\begin{abstract}

Fake review identification is an important topic and has gained the interest of experts all around the world. Identifying fake reviews is challenging for researchers, and there are several primary challenges to fake review detection. We propose developing an initial research paper for investigating fake reviews by using sentiment analysis. Ten research papers are identified that show fake reviews, and they discuss currently available solutions for predicting or detecting fake reviews. They also show the distribution of fake and truthful reviews through the analysis of sentiment. We summarize and compare previous studies related to fake reviews. We highlight the most significant challenges in the sentiment evaluation process and demonstrate that there is a significant impact on sentiment scores used to identify fake feedback.

\end{abstract}


\section{Introduction}
Online purchasers on e-commerce sites are increasing day by day. Online purchasers often post reviews about specific products they have used. The importance of user reviews can be viewed from the user and business perspectives. From the user's perspective, these reviews can influence new customers to decide on a certain product in a good or bad way. For purchasing online, users often visit e-commerce sites rich with user experience about products. So quality and number of user experiences can affect user traffic on site. The fake review detection task is one of the challenging classification tasks in the field of knowledge discovery. Multiple angles of capturing deception in review data have been focused on by researchers for a decade. The focus of our research work is to investigate the techniques and classification model to identify individual fake reviews by analyzing different perspectives of review data. Sentiment Analysis (SA) is the branch of Natural Language Processing (NLP) in charge of the design and implementation of models, methods, and techniques to determine whether a text deals with objective or subjective information and, in the latter case, to determine if such information is expressed in a positive, neutral, or negative way as well as if it is said strongly or weakly. Sentiment analysis is a technique that combines machine learning and natural language processing (NLP) to achieve the goal. Since a large part of the subjective content expressed by users on social networks is about opinions (on review sites, forums, message boards, chats, etc.), SA is also known as Opinion Mining (OM). The expression of sentiment plays an important role in fake news. Social media users tend to comment on posts when there is content that they consider arousing but over which they feel less in control. Conversely, users tend to share a post when they feel more in control. Most people prefer buying a product having maximum positive feedback or a 5-star rating. The detection of fake reviews has become a hot research issue. Despite the efforts of existing studies on fake review detection, the issues of imbalanced data and feature pruning still lack sufficient attention. Online fake reviews are reviews that are written by someone who has not used the product or the services. Fake online reviews in e-commerce significantly affect online consumers, merchants, and, as a result, market efficiency \cite{r1}.

\section{Related work}
\label{sec:headings}
\paragraph{Shetgaonkar et al., \cite{r2}} This research paper use NLP and SA methods to identify fake reviews and feedback. This paper transferred the data to be able to analyze and detect the review's feedback. This paper proposes a system that detects fake reviews as shown in Fig. 1. The data set used in this paper has 127,152 reviews of which 13,152 are positive reviews and 13,279 are negative reviews. So, a pre-processing technique is text cleaning, an implemented method involving analyzing review feedback with deep learning neural networks like the Gated GRU, LSTM, and Bi-LSTM. The result evaluation was done by using various activation functions like ReLu, TanH, and Sigmoid as well as other hyperparameters like dropout, and changing the layers in the model was carried out to get the best model configurations. This paper's data size is not enough. In the future, they will increase the size of the data set used.

\begin{figure}[ht]
  \centering
  \includegraphics[width=0.4\linewidth]{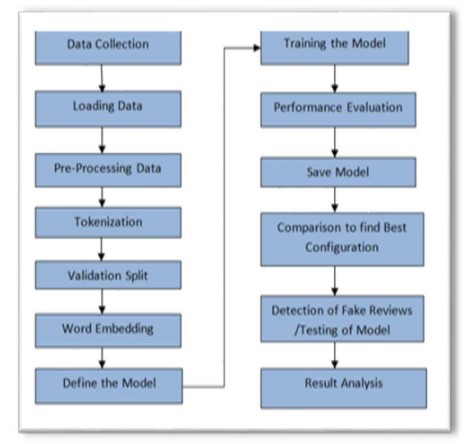}
  \caption{The Proposed Methodology}
  \label{fig: Fig1}
\end{figure}

\paragraph{Peng et al., \cite{r3}} This work compared various sentiment lexicons and discovered that MPQA+Product had the best accuracy of 61.4 percent. When various approaches were compared to different datasets, the sentiment score approach proved to be the most accurate. In this paper, the goal is to incorporate sentiment analysis into spam review detection. The proposed method computes sentiment score from the natural language text by a shallow dependency parser and generates a time series combined with discriminative rules for efficient detection of the spam score and spam reviews were proposed that include sentiment analysis techniques into review spam detection. This research has three tasks, the first task is to generate a sentiment lexicon and compute the sentiment score.  The second task is to set up a set of discriminate rules and, the third task is to establish a time series method to detect spam reviews. This paper proposed different algorithms for the tasks mentioned above.

\paragraph{Elmurngi et al., \cite{r4}} Sentiment Analysis (SA) has become one of the most interesting topics in text analysis. This paper aims to classify movie reviews into 2 groups of positive or negative polarity by using machine learning algorithms. Moreover, SA and text classification methods are applied to a dataset of movie reviews. This study proposed several methods to analyze a dataset of movie reviews. This paper used five techniques to detect fake reviews and they are Na¨ıve Bayes (NB), Support Vector Machine (SVM), K-Nearest Neighbors (KNN-IBK), KStar (K*), and Decision Tree (DT-J48 as shown in Fig. 2. The results of the experiments show that the SVM algorithm is the most accurate one. This study needs to use different datasets such as Amazon or eBay, and also use different methods like Python.

\begin{figure}[ht]
  \includegraphics[width=0.5\linewidth]{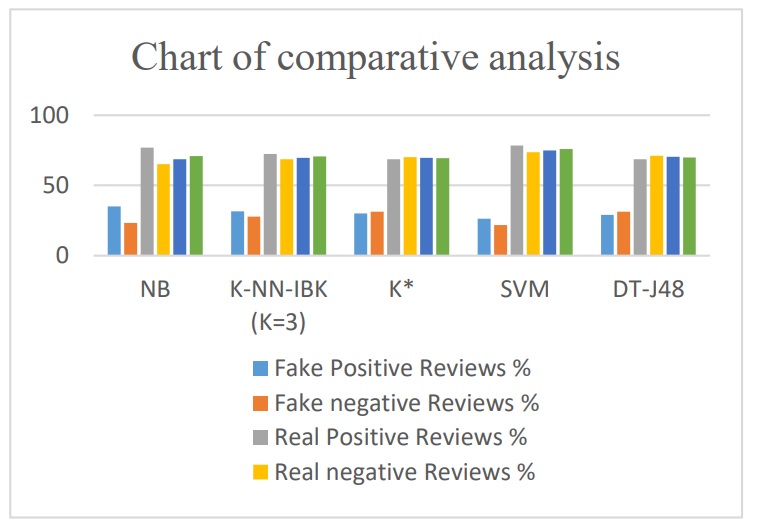}
  \centering
  \caption{Comparative analysis of all methods.}
  \label{fig: Fig2}
\end{figure}

\paragraph{Devika. et al., \cite{r5}} Sentiment analysis is a field of study that analyzes people's sentiments, feelings, or emotions towards bound entities. This paper tackles an elementary downside of sentiment analysis, sentiment polarity categorization. The researcher used some supervised and semi-supervised algorithms used and supported Vector Machine (SVM). In light of this study, they depend on software systems that can facilitate the user to get the correct product. This software system can do analysis and so if any pretend review is found from any information processing address systematically then the admin user will block that information processing address. Fig3 represents the System Architecture.

\begin{figure}[ht]
  \centering
  \includegraphics[width=0.6\linewidth]{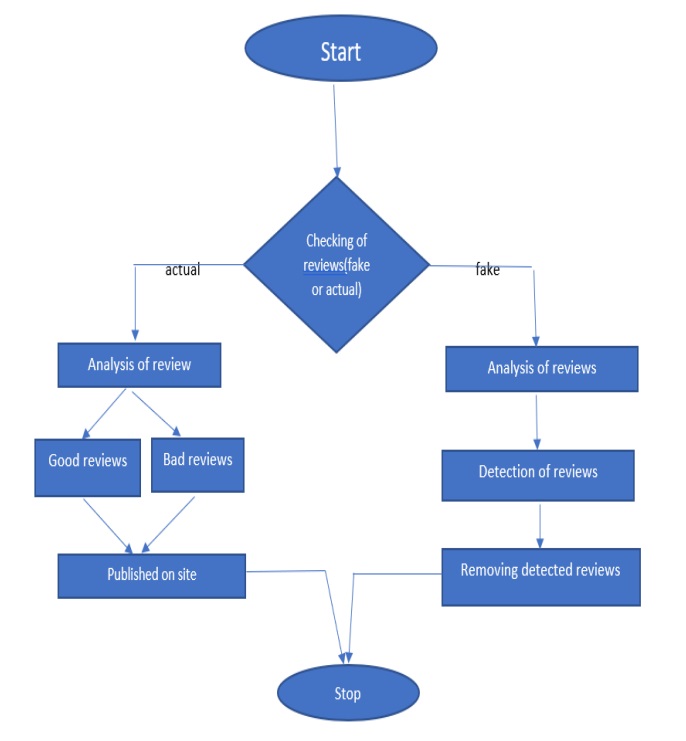}
  \caption{System Architecture}
  \label{fig: Fig3}
\end{figure}

\paragraph{Kashti. et al., \cite{r6}}  Online shopping is increasing day by day as every product and service is getting available easily. This paper presents an active learning method for detecting fake reviews conducted on real-life data. Detecting fake reviews has become a more important issue for customers to make better decisions on purchases as well as for the trader to make their products reliable. This study develops a System to accept Reviews from Authenticated Users only and will be using an NLP-based sentiment analyzer and text mining algorithms to classify and predict positive, negative, and neutral reviews. This paper suggests different factors for identifying fake reviews. Furthermore, this study was able to detect fake positive reviews and fake negative reviews through detection processes.

\paragraph{Chang. et al., \cite{r7}} Customers can post their reviews or opinions on several websites. These reviews are helpful for organizations and for future consumers, who get an idea about products or services before selecting them. This paper uses sentimental analysis to detect fake comments and calculates the distance between different positive sentiments or negative sentiments through emotional analysis. This study uses the textual data 1930 data about food and commodities from Amazon. This paper starts with identifying fake comments by using sentimental analysis then comparing them with the previous ways and using a clustering algorithm, after collecting the score to evaluate the sentiment of the comment they use the package, textbook they used CFSFDP clustering algorithm to identify fake comments then remove it. In the end, they perform iterative clustering to ensure the accuracy of the results.

\paragraph{Wang. et al., \cite{r8}}  In this paper the author mentioned that most existing methods have lower accuracy in identifying fake reviews because they just use single features and lack categorized experimental data. Therefore, it is necessary to detect and filter fake reviews. To solve this problem, a method to detect fake reviews based on multiple feature fusion and rolling collaborative training. First, this method involves an initial index system with multiple features such as text features, behavior features of critics, and sentiment features of reviews. Then the method needs an initial training sample set. So the related algorithms are designed to extract all the features of a review. Finally, the method uses the initial sample set to train 7 classifiers, and the most accurate one will be selected to classify new reviews. In the future, they will strive to find a more effective and accurate detection method that can detect false information in multiple fields, including fake information, fake news, and rumors.

\paragraph{Elmogy. et al., \cite{r9}} This paper compares different sentiment classification algorithms. They apply the sentiment classification algorithms using two different datasets with stop words which is the more efficient way of text categorization and detection of fake reviews. In this paper, they were able to detect fake positive and negative reviews through detection processes. This research paper used five supervised learning algorithms to classify the sentiment of the datasets: Naïve Bayes (NB), Support Vector Machine (SVM), K-Nearest Neighbors (KNN-IBK), KStar (K*), and Decision Tree (DT-J48) as shown in Figure[4]. They found that the SVM algorithm is the most accurate method to detect fake positive and negative reviews.

\begin{figure}[ht]
  \centering
  \includegraphics[width=0.5\linewidth]{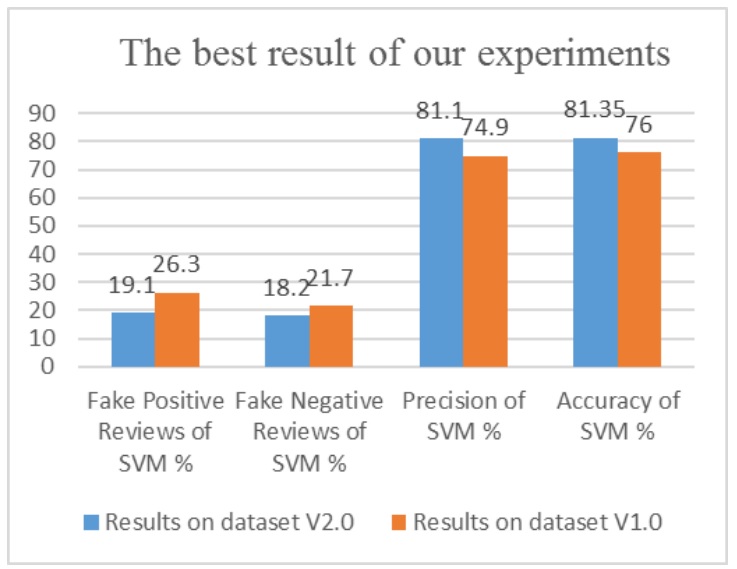}
  \caption{five supervise learning algorithms classifying the sentiment of the datasets}
  \label{fig: Fig4}
\end{figure}

\paragraph{Fang. et al., \cite{r10}}  Sentiment analysis seems to have a strong foundation with the support of massive online data. This paper tackles a fundamental problem of sentiment analysis, sentiment polarity categorization. Data used in this paper is a set of product reviews collected from amazon.com. for this study, they used a software called sci-kit-learn, which is an open-source ML software in Payton and the classification models are Naïve Bayesian, Random Forest, and Support Vector Machine. They aim to tackle the problem of sentiment polarity categorization, which is one of the fundamental problems of sentiment analysis.

\paragraph{Poonguzhali. et al., \cite{r11}} Users directly take decisions based on reviews or opinions that are written by others based on their experiences. Researchers analyzed the reviews posted for the products, using the concept of Sentiment Analysis with supervised machine learning. This paper uses the SVM algorithm to detect fake reviews of the products and classify the reviews into positive and negative. It developed an online interface by having a login so users can easily find the products and find higher quality products based on genuine reviews and eliminate Negative reviews after analyzing ratings, reviews, and smileys.

\begin{figure}[ht]
  \centering
  \includegraphics[width=0.6\linewidth]{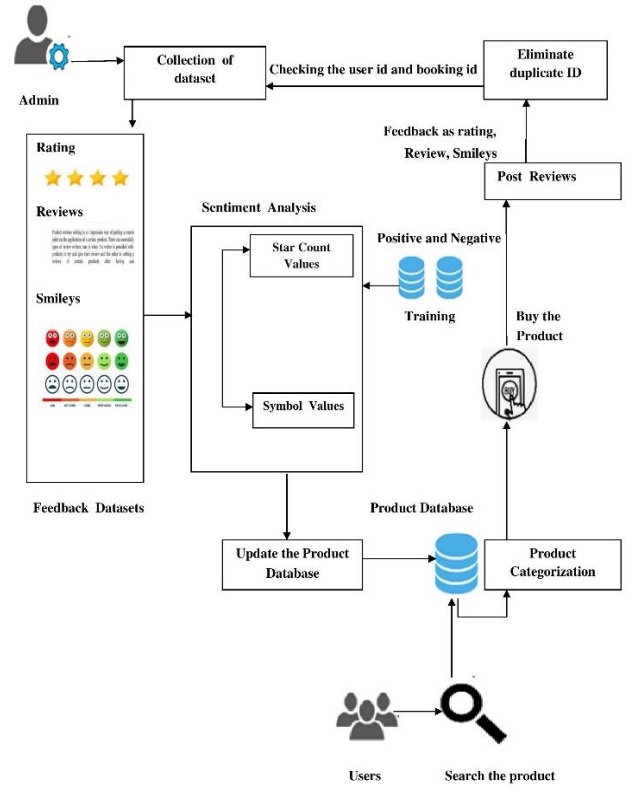}
  \caption{\ Architecture}
  \label{fig: Fig5}
\end{figure}

Figure 5 shows customers can give reviews only once. So, this will prevent Fake Reviews in this proposed method. So, only purchased users can post the reviews, and duplicates are verified based on user id and booking id Genuine reviews are considered for product recommendation.

\section{Future work}
Fake review detection is considered an important problem for reflecting genuine and legitimate user experiences and opinions. Machine learning techniques were highly used in the automatic detection of fake reviews and solving this problem. The dataset used in most studies only contains data from two Yelp and Amazon datasets. Also, for future research, may intend to evaluate the models with more various datasets. Using other datasets such as eBay and using different feature selection methods. Consider applying sentiment classification algorithms to detect fake reviews such as Python, Statistical Analysis System (SAS), and Stata. in addition, increasing the size of the dataset used gives more accurate training of the model and better results.

\section{Conclusion}
Nowadays the usage of the Internet and online marketing has become very popular. Millions of products and services are available in online marketing that generates a huge amount of information. This paper reviews the literature on Fake Review Detection (FRD) on online platforms. It presented a systematic literature review of the spam review detection domain and highlighted recent research contributions in the form of different feature engineering approaches, spam review detection methods, and different measures used for performance evaluation. Various sentiment analysis methods are reviewed in this work by combining NLP and ML techniques. From the literature survey conducted on the number of research papers for sentiment analysis on various kinds of formats, we found that the most commonly used Machine Learning techniques were Naïve Bayes and SVM classifiers. Moreover, sentiment analysis has been used in this work to determine the customer's opinions through online feedback given by them on products.

\bibliographystyle{plain}  
\bibliography{references.bib}  

\end{document}